\documentclass{article}

\usepackage{amssymb}
\usepackage{amsmath}
\usepackage{PRIMEarxiv}
\usepackage{subcaption}
\usepackage{dsfont}
\usepackage[utf8]{inputenc} 
\usepackage[T1]{fontenc}   
\usepackage{hyperref}      
\usepackage{xurl} 
\usepackage{booktabs}      
\usepackage{amsfonts}      
\usepackage{nicefrac}       
\usepackage{microtype}     
\usepackage{lipsum}
\usepackage{fancyhdr}      
\usepackage{graphicx}      
\usepackage{algorithmic}
\usepackage{romannum}

\title{Adapting a Foundation Model for Lunar Surface Height Estimation
}

\author{
\textbf{Patrick Bauer} \\
University of Technology of Troyes \\
Hochschule Darmstadt \\
EUT+ Data Science Institute\\
\texttt{patrick.bauer@utt.fr}
\And
\textbf{Marius Schwinning} \\
European Space Agency\\
\texttt{marius.schwinning@esa.int}
\And
\textbf{Melanie Siegel} \\
Hochschule Darmstadt\\
EUT+ Data Science Institute\\
\texttt{melanie.siegel@h-da.de}
\And
\textbf{Andreas Weinmann} \\
Technische Hochschule Würzburg-Schweinfurt \\
\texttt{andreas.weinmann@thws.de}
\And
\textbf{Hichem Snoussi} \\
University of Technology of Troyes \\
EUT+ Data Science Institute\\
\texttt{hichem.snoussi@utt.fr}
}

\begin{document}

\noindent{\footnotesize\copyright~2026 IEEE. Personal use of this material is permitted.
Permission from IEEE must be obtained for all other uses, in any current or future media,
including reprinting/republishing this material for advertising or promotional purposes,
creating new collective works, for resale or redistribution to servers or lists, or reuse
of any copyrighted component of this work in other works.\par}
\vspace{1em}

\vspace{0.5em}

\noindent{\footnotesize
Accepted for publication in \textit{IEEE Geoscience and Remote Sensing Letters}.
DOI: \href{https://doi.org/10.1109/LGRS.2026.3754087}{10.1109/LGRS.2026.3754087}\par}

\vspace{1em}
\maketitle

\begin{abstract}
Digital elevation models (DEMs) can provide accurate height information, making it invaluable for analyzing the lunar surface. Due to recent advancements in computer vision, the focus is on generating DEMs using monocular depth estimation via deep learning. Classic lunar crater detection methods rely mainly on 2D image data. Our goal is to address this issue by developing a relative lunar surface height estimator that could provide additional information for crater localization. In this letter, we present a methodology that builds on the  zero-shot relative depth estimation model Depth Anything V2 (DAV2). Other works have been using it as a state-of-the-art comparison for their proposed lunar DEM estimation method, but without adaptations to the target domain. Thus, it may underperform. Therefore, we propose a fine-tuning strategy with publicly available SPG-derived DEM data of the lunar surface, and refer to it as DEM-DAV2. Specifically, we introduce new learnable parameters in the Transformer-based encoder with low-rank adaptation (LoRA), and we unfreeze certain layers in the model's decoder. On our test set, the proposed DEM-DAV2 reduces the average MAE and RMSE by $\sim$80\% and $\sim$79\%, respectively, compared to the zero-shot DAV2. This demonstrates a significant improvement in performance compared to the zero-shot model, effectively transforming DAV2 into a reliable relative height estimator of the lunar surface. However, the predictions remain relative and require scale recovery from a low-resolution DEM such as SLDEM.
\end{abstract}

\section{Introduction}
In recent years, researchers have developed promising approaches to detect lunar craters using deep learning techniques to determine their locations and distribution~\cite{tewari2022,zou2024,Bauer2026}. These approaches typically involve lunar image data instead of digital elevation models (DEMs). By solely relying on image data, one is leaving out the third dimension and this can be fatal for detection quality, as some degraded craters are hard to detect~\cite{Tewari2023DeepLB}. The quality of crater detection could benefit by incorporating height information. Geologists and lunar researchers are already highly interested in accessing the most accurate and highest-resolution DEMs possible. In general, DEM generation can be approached in various ways.  Techniques such as Shape from Shading (SfS)~\cite{Alexandrov2018} and stereophotogrammetry (SPG)~\cite{Henriksen2017} can be applied to lunar images to generate DEMs. SPG, that typically need at least two images of the same scene with disparate emission angles, generates DEM products with a spatial resolution of up to 2 m/px but faces significant limitations since stereo observations covering the same scene are very limited. According to Wagner et al.~\cite{Wagner_2024}, approximately 5\% of the lunar surface is covered by stereo pairs. 

Deep learning-based approaches have been very prominent in various computer vision tasks. In the field of monocular depth estimation (MDE), various deep learning based models were introduced. These models can estimate depth information without requiring multiple images of the same scene. Models such as MiDAS~\cite{Ranftl2019TowardsRM}, Marigold~\cite{Ke2023RepurposingDI}, Depth Anything~\cite{Yang2024DepthAU} and its successor Depth Anything V2 (DAV2)~\cite{Yang2024DepthAV} yield superior MDE results. They have the additional advantage of being foundation models. Foundation models are a class of models that have been extensively pre-trained on vast amounts of data and can be used for various downstream tasks. They are often based on a Transformer architecture~\cite{Vaswani2017AttentionIA} and, therefore, on attention modules. Recently, they have been applied more frequently in the area of remote sensing~\cite{foundation_rs}. However, naively applying foundation models to remote sensing tasks does not yield good results due to a significant domain gap between natural scenery images in pre-training and remote sensing data in the downstream tasks, cf. Luo et al.~\cite{samrsis}. Therefore, adapting and fine-tuning are required.

In the field of deep learning-based MDE for planetary surfaces such as the Moon and Mars, a broad range of approaches and techniques exist. Several works~\cite{Liu2022, ChenTianhao2025, tao2021,Chen2022,LAGRASSA2026100118, Chen2024} have been introduced and yielded very promising and highly accurate models trained for that task. For example, Chen et al. proposed LDEMGAN2.0~\cite{ChenTianhao2025}, a generative adversarial network (GAN) processing a narrow angle camera (NAC) image and the coarse SLDEM as a dual-channel input for absolute lunar height estimation, extending LDEMGAN1.0~\cite{Liu2022}. The performance notably increased with the implementation of a self-similarity loss compared to its predecessor. Tao et al.~\cite{tao2021} introduced MADNet, based on a GAN, trained on HiRISE PDS orthorectified images (ORIs) and photogrammetric DEMs of the Martian surface. MADNet outperforms SPG and SfS in both flexibility and speed. La Grassa et al.~\cite{LAGRASSA2026100118} proposed the Transformer-based SurfPlaNet network for Martian DEM estimation, taking a coarse DEM as an auxiliary input channel. This network is capable of predicting scale-aware DEMs, remaining tied to the global elevation prior. Chen et al.~\cite{Chen2024} developed ELunarDTMNet, which comprises a dual-branch encoder, using hierarchical Transformer blocks for images and convolutional blocks for DEMs.  This is followed by a fusion module and a residual-connected decoder module. The objective is to estimate the absolute height of the lunar surface, which is generally the aim of most proposed methods. Notably, the aforementioned works train task-specific models, rather than building on a pre-trained depth foundation model. To the best of our knowledge, the parameter-efficient adaptation of foundation models for orbital MDE remains unexplored. La Grassa et al.~\cite{LAGRASSA2026100118} and Osadnik et al.~\cite{osadnik2025} compared their approach with a state-of-the-art (SoTA) model, DAV2. However, due to the aforementioned domain gap, we argue that the comparison with a zero-shot model is limited.
The aim of this letter is to determine if the DAV2 model can be effectively adapted for estimating relative lunar surface height. Therefore, our objectives are:
\begin{enumerate}
        \item [(i)]To propose a method to adapt the DAV2 model by inserting new trainable parameters with LoRA and unfreezing certain layers in the decoder.
        \item [(ii)] To create a training set and a test set from the SPG-derived, high-resolution DEM products by splitting them by geocoordinates so that no terrain is shared between the two sets.
        \item [(iii)] To evaluate the improvement gain of DEM-DAV2 compared with DAV2 quantitatively and qualitatively.
\end{enumerate}
This enables us to develop an auxiliary model that could support existing crater detectors, which currently mainly consider 2D data.

\section{Data and Methodology}
\subsection{Available Lunar DEM Data}
NASA launched the Lunar Reconnaissance Orbiter (LRO) in 2009, capturing high-resolution images of the Moon~\cite{Robinson2010}. Taking images of the same scenery on subsequent orbits from an oblique angle allows researchers to create SPG-derived DEMs of the lunar surface with a resolution of up to 2 m/px~\cite{Henriksen2017}. This approach relies on acquiring data on subsequent orbits. However, available stereo pairs only cover around 5\% of the lunar surface~\cite{Wagner_2024}. DEM products based on the Lunar Reconnaissance Orbiter Laser Altimeter (LOLA) produce very accurate height maps, albeit at the cost of coarse resolution. For example, the SLDEM~\cite{Barker2015ANL} has a resolution of 60 m/px. In order to be able to reconstruct the elevation profile of the lunar surface containing high-frequency details, we select the available SPG-derived DEM data products with a resolution of 2 m/px and their corresponding orthorectified images (ORIs) and download the data from \url{https://data.lroc.im-ldi.com/lroc/rdr_product_select}. To generate a train-test-validation split, we first crop all the selected ORIs and DEMs into tiles of size 259$\times$259. To prevent data leakage, we ensure that the same physical region on the Moon is strictly assigned to one of three sets. Since some DEMs contain \textit{nodata} values, we follow the strategy outlined in Tao et al.~\cite{tao2021}, where tiles with \textit{nodata} values are removed. In total, we obtained 84,853 training images, 10,920 validation images, and 13,020 test images to align with an approximate 80—10—10 split. The corresponding images were acquired at solar incidence angles of 26--72$^\circ$ and cover latitudes from 73$^\circ$S to 63$^\circ$N.

\subsection{Depth Anything V2}

Yang et al.~\cite{Yang2024DepthAU} introduced a MDE model called Depth Anything. Due to its architecture and the vast amount of training data, it is capable of zero-shot depth estimation on natural images. Subsequently, Yang et al. developed the successor, Depth Anything V2 (DAV2)~\cite{Yang2024DepthAV}, incorporating real pseudo-labeled images into training process.  We use the ViT-L variant of DAV2 throughout. DAV2 consists of a DINOv2~\cite{Oquab2023DINOv2LR} based encoder, by stacking 24 Transformer blocks. The decoder utilizes the DPT\cite{Ranftl2021VisionTF} decoder system, which is a convolutional neural network (CNN)-based model. It consists of three-stage Reassemble operations and the core in the decoder are RefineNet-based feature fusion blocks~\cite{Lin2016RefineNetMR, Xian2018MonocularRD}. After the processing through the Transformer modules, 4 tensors at different stages serve as the inputs to the decoder. The decoder output is the inverse depth map where the values represent relative inverse depth. During pre-training of DAV2, a scale-and-shift invariant (SSI) loss was minimized based on Ranftl et al.~\cite{Ranftl2019TowardsRM}. To develop a zero-shot capable MDE model, they combined various ground truth (GT) datasets  with different scales and shifts. An SSI-based loss function overcomes this challenge. Precisely, let $d$ and $\hat d$ be the ground truth and the predicted depth map, respectively, both with height $H$ and width $W$. Before minimizing the loss function, the normalization
\begin{equation}
        d^* = \frac{d - m(d)}{s(d)},\quad \hat{d}^* = \frac{\hat{d} - m(\hat{d})}{s(\hat{d})}~\label{eq: norm},
\end{equation}
is applied, where $m(d)$ denotes the median of $d$ and $s(d)$ is the mean absolute deviation from $m(d)$.
Each map is normalized using its own median and scale. We adopt the same normalization for training our model.
\subsection{Finetuning Strategy}
Fine-tuning foundation models and applying it for downstream tasks is an increasing popular approach in machine learning. It is a non-trivial task as naively fully-unfreezing the model and update all parameters is inefficient~\cite{LoRA}. As a consequence, researchers focused on developing parameter efficient fine-tuning strategies (PEFT) such as low-rank adaptation (LoRA)~\cite{LoRA}. It is a widely used and popular method to update parameters using LoRA in Transformer modules due to their intrinsically low-rank structure of weight updates during fine-tuning. More precisely,  low-rank matrices are introduced and added to the existing high-dimensional matrices. 

The background behind is that the update steps of neural networks even for high dimensional matrices essentially happens in low-rank subspaces~\cite{LoRA}. Further, it has the advantage that it does not add any inference latency. Considering its advantages, we choose to update the Transformer-based encoder with LoRA by applying it to the query and value matrices with rank $r = 8$. At the beginning of the training, the low-rank update is initialized to zero, so the training begins with the pretrained model's parameters.

Instead of fully-unfreezing the CNN-based decoder network, we selectively unfreeze certain layers. Therefore, we keep the first layers of the decoder, namely the \textit{Reassemble} layers, frozen. We unfreeze 3 of 4 \textit{RefineNet} blocks and the output head. Specifically, we keep the \textit{RefineNet} block 4 frozen since, as the deepest block, it only receives a single input, meaning that only one of its two residual units is active during the forward pass. This also prevents a further increase of the imbalance between the number of learnable parameters in the decoder and the encoder.  In total, we fine-tune $\sim 8$ million parameters out of $\sim 335$ million.

We choose a hybrid loss function $L_\text{total}$ that consists of three partial losses
\begin{equation}
L_\text{total} = \alpha L_{\text{Berhu}} + \beta L_{\text{gm}} + \gamma
L_\text{norm},~\label{eq: loss}
\end{equation} 
with hyperparameters $\alpha, \beta, \gamma$ and $L_{\text{Berhu}}, L_{\text{gm}}, L_\text{norm}$ as explained next.
We base our approach on well-established loss functions for DEM estimation in previous studies. For example, Tao et al.~\cite{tao2021}  employed the Berhu~\cite{Zwald2012TheBP} and a gradient matching loss term, and Chen et al.~\cite{Chen2022} incorporated a hybrid loss function containing a normal loss term. We apply Eq.~\ref{eq: norm} to both the prediction and the GT before calculating the loss function. The Berhu loss is defined as
\begin{equation}
    L_\text{Berhu}(d^*, \hat{d}^*) = \frac{1}{HW}\sum_{i=1}^{HW}
    r_i\, \mathds{1}_{\{r_i \leq \tau\}}
    + \frac{r_i^2 + \tau^2}{2\tau}\, \mathds{1}_{\{r_i > \tau\}},
\end{equation}
where $r_i = \lvert d_i^* - \hat{d}_i^* \rvert$ and $\tau = \frac{1}{5}\max_j r_j$.

The multiscale gradient matching loss $L_{\text{gm}}$ accounts for the deviation of the residual gradient of the prediction with respect to its corresponding GT. It incorporates downscale operations, that have been proven highly effective in~\cite{Ranftl2019TowardsRM}.
\begin{equation}
        L_{\text{gm}}(d^*, \hat{d}^*) = \frac{1}{HW} \sum_{k=1}^{K}\sum_{i = 1}^{HW}  \lvert \nabla_x R_i^{(k)} \rvert + \lvert \nabla_y R_i^{(k)}\rvert,
\end{equation} 
where $R_i =  d_i^*-\hat{d}_i^*$ and $R^{(k)}$ denotes the disparity of the corresponding maps at scale $k$, where we set $K=4$. The image resolution is halved at each scale level.

The third part is a normal surface loss $L_{\text{norm}}$ that measures the accuracy of the normal to the surface of the prediction and its corresponding GT. 
\begin{equation}
        L_{\text{norm}} = \frac{1}{HW}\sum_{i=1}^{HW} \bigl (1- \frac{\langle n_{d_i^*}, n_{\hat{d_i^*}} \rangle}{{\Vert  n_{d_i^*}\Vert \Vert n_{\hat{d_i^*}} \Vert}}\bigr ) .
\end{equation}
The normal vector is given as  $n = (-\frac{\partial d^*}{\partial x}, -\frac{\partial d^*}{\partial y}, 1)^T$. By applying the normalization to the prediction and the GT with Eq.~\eqref{eq: norm}, the infinitesimal changes in $x$ and $y$ direction are only a small fraction compared to the third component of the normal. As a consequence, this term would add only marginally to the overall loss. Therefore, we apply a scale $j$ to the first two components of the normal  and empirically choose $j = 30$ as a vertical scaling of the normalized heights. Note that $j$ does not undo the normalization in Eq.~\eqref{eq: norm}, since $s(d)$ differs for each tile. It is chosen such that the scaled slopes are of the order of the third component. Henceforth, we refer to the fine-tuned DAV2 model as DEM-DAV2. We empirically choose $\alpha = 1.0$, $\beta = 2.0$ and $\gamma = 0.5$.

\subsection{Implementation Details}

To make the training procedure more robust and produce more artificial solar azimuth conditions, we apply basic data augmentation strategies. Specifically, we apply flipping in the vertical and horizontal directions, both with a probability of 0.5. In total, we train for 20 epochs with a per-GPU batch size of 8 and train parallel on 4 H100 GPUs with 96 GB RAM each. We apply weight decay of 1 $\times$ 10$^{-2}$ and use the AdamW~\cite{adamW} optimizer. The learning rate for the encoder and decoder are chosen differently, both utilize a linear warmup for the first 10\% of the iterations, and are then decayed to zero at full training with a cosine schedule. The initial learning rate for the LoRA weights is set to 1$\times$10$^{-4}$ and the initial learning rate for the decoder to 5$\times$10$^{-5}$. We conduct experiments with the model that has the lowest validation loss (Eq.~\ref{eq: loss}), which was reached at epoch 13.

\section{Results}

We show that DAV2 can be transformed into a relative depth estimator for the lunar surface. It was used as a state-of-the-art (SoTA) comparison in the work of La Grassa et al.~\cite{LAGRASSA2026100118} and Osadnik et al.~\cite{osadnik2025}. We demonstrate improved performance by fine-tuning it on lunar DEM data to create a promising lunar DEM estimator. We  compare the results of DAV2 with those of our DEM-DAV2 on our test set.  

The test set consists of 13,020 images of size 259$\times$259. As we minimized an scale-and-shift invariant (SSI) loss, both the predictions of DAV2 and DEM-DAV2 are defined up to an affine
transformation. To evaluate in absolute metrics, we follow Osadnik et
al.~\cite{osadnik2025} and align each prediction $\hat{d}\in\mathbb{R}^{259\times259}$ with the corresponding GT elevation map $d\in\mathbb{R}^{259\times259}$ (in meters) by
\begin{equation}
        \hat{d}^+ = \sigma_{\text{GT}}\cdot \frac{\hat{d} - \mu_{\hat{d}}}{\sigma_{\hat{d}}} + \mu_{\text{GT}},~\label{eq: alignment}
\end{equation}
with the corresponding mean values $\mu_{\hat{d}}, \mu_{\text{GT}}$ and the standard deviations $\sigma_{\text{GT}}$ and  $\sigma_{\hat{d}}$.
To evaluate the models' performances quantitatively, we compute the mean absolute error (MAE) and the root mean squared error (RMSE) between $d$ and $\hat{d}^+$
\begin{align}
        \text{MAE} &= \frac{1}{259^2}\sum_{i = 1}^{259^2}\vert d_i - \hat{d_i}^+ \vert, \\
        \text{RMSE} &= \sqrt{\frac{1}{259^2}\sum_{i = 1}^{259^2}\bigl ( d_i - \hat{d_i}^+ \bigr )^2 }.
\end{align}
\begin{figure}[ht!]
    \centering
    \includegraphics[width=0.75\textwidth]{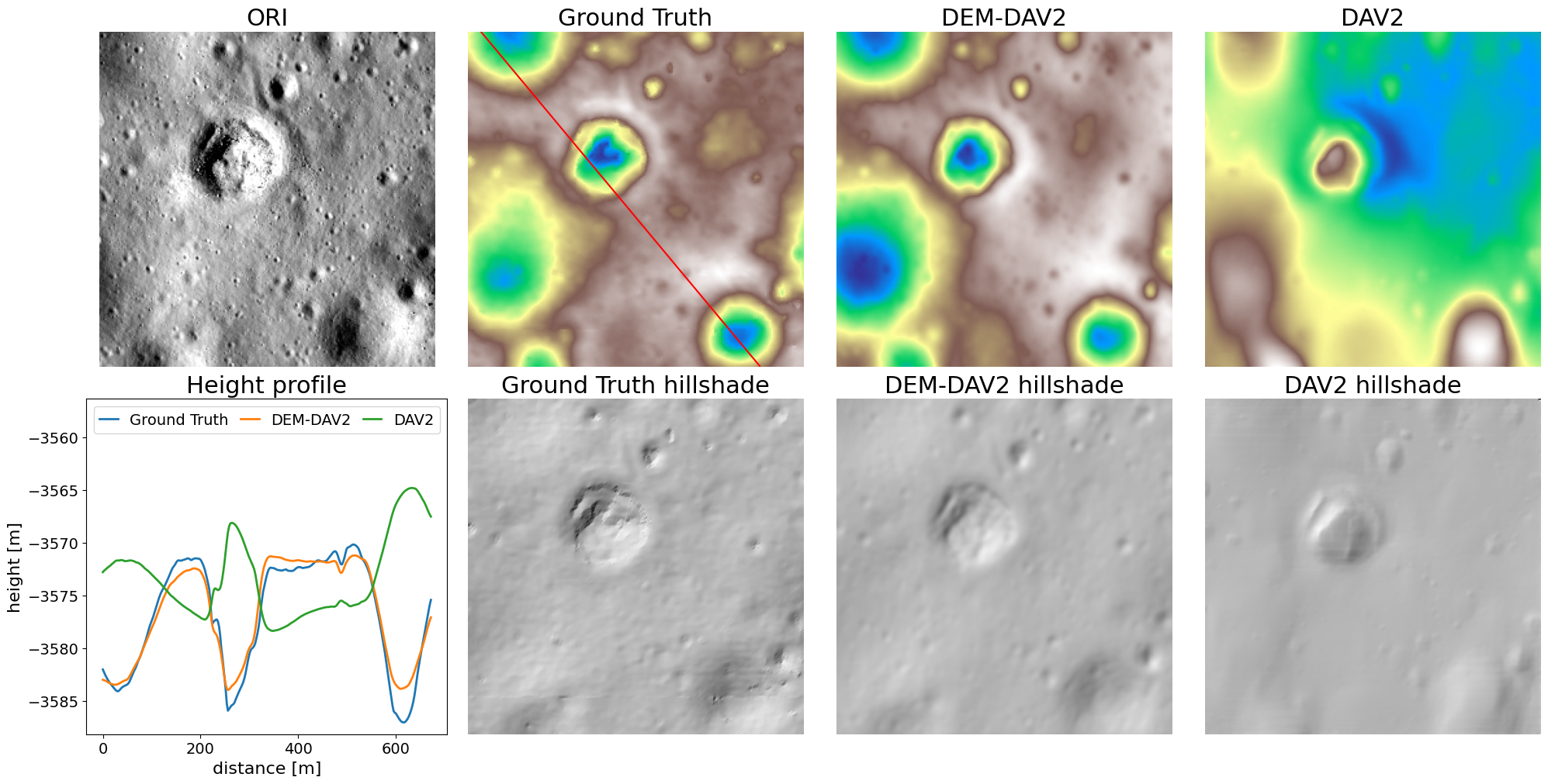}\\
    \includegraphics[width=0.75\textwidth]{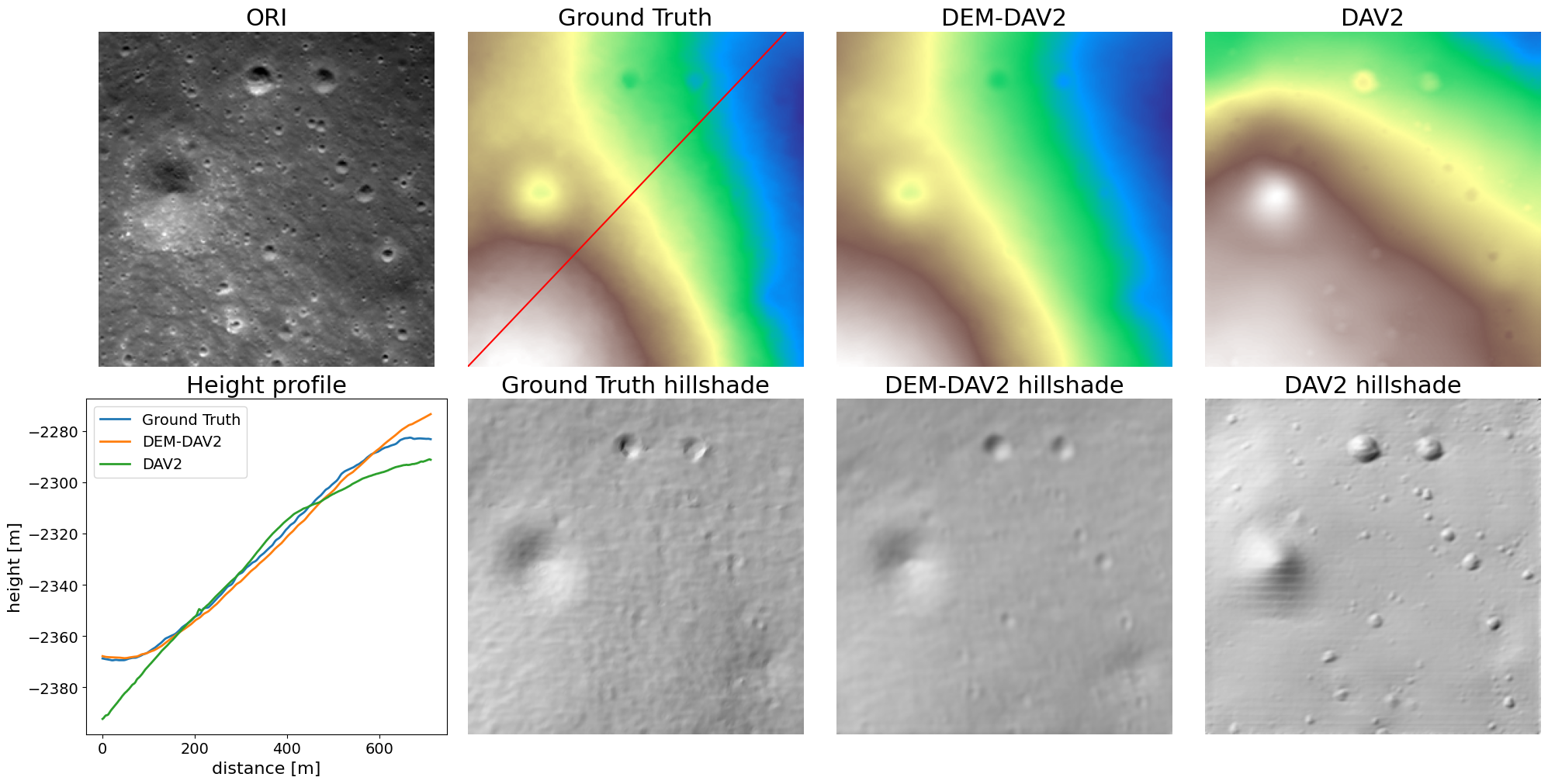}\\
    \includegraphics[width=0.75\textwidth]{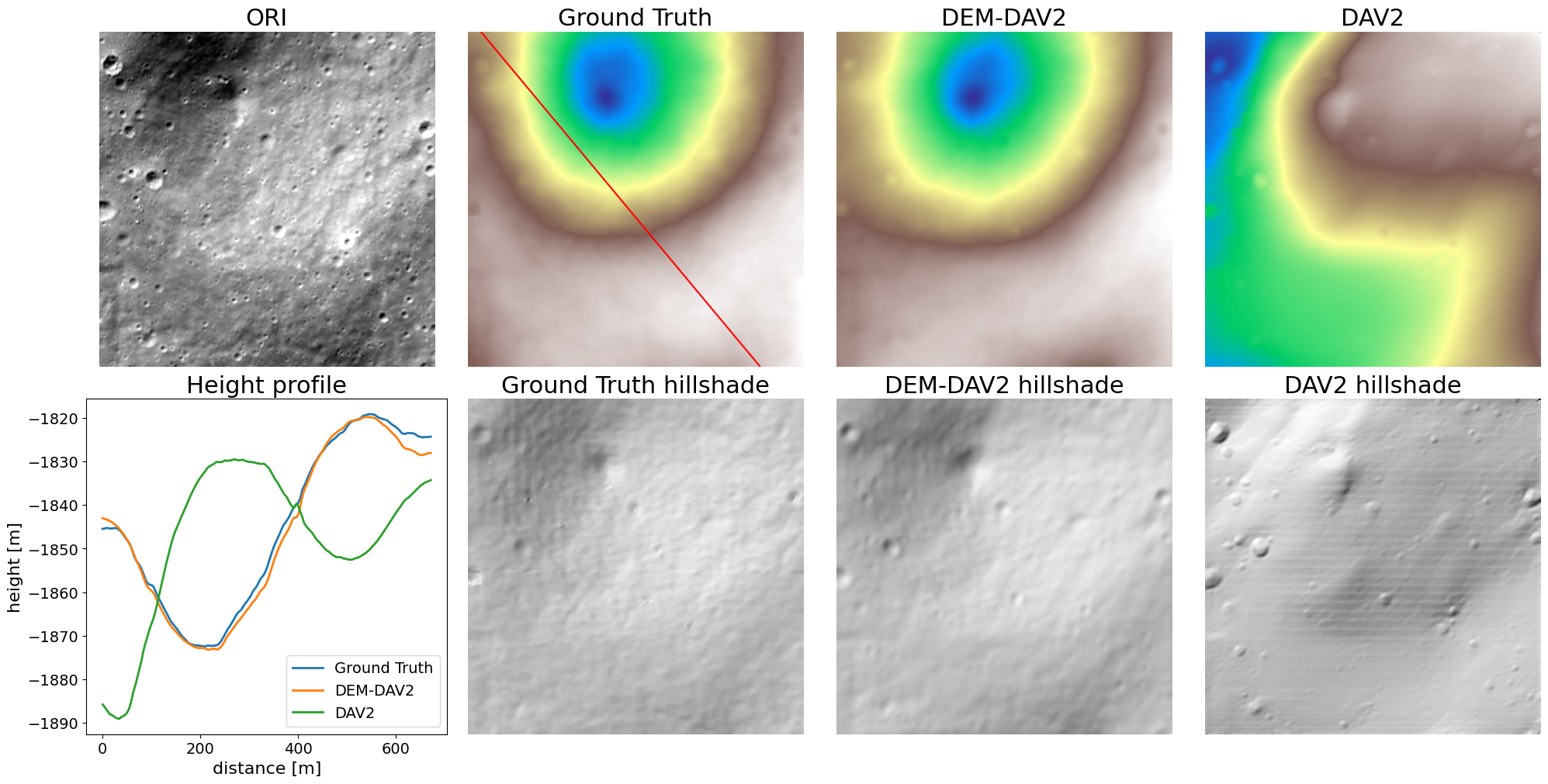}\\
    \caption{We illustrate the corresponding predictions from DAV2 and DEM-DAV2 for three example ORI images after aligning them with the ground truth (GT) belonging to the ORI. We observe a significant improvement in DEM estimation quality using the proposed approach (first row of each example). The second row confirms this improvement with profile analysis and hillshades.}~\label{fig:comparisonzeroshotandfinetuned}
\end{figure}
\subsection{Qualitative Evaluation}
First, we evaluate the performance of DEM-DAV2 by comparing it visually on examples of our test dataset with the zero-shot performance of DAV2. Fig.~\ref{fig:comparisonzeroshotandfinetuned} illustrates a strong performance improvement of DEM-DAV2 compared to DAV2. While DAV2 captures the location and outline of local high-frequency features, such as craters, it often reverses their relief and fails to capture the overall relief and structural elevation differences. Here, DEM-DAV2 has a significant advantage, because it better captures high-frequency features and excels in determining the relief. We determine the hillshades for the example images for further analysis. Notably, the DEM-DAV2 hillshades show a clear improvement compared to the DAV2 hillshades, as DAV2 frequently inverts the crater relief, predicting domes instead of craters. It is evident that the alignment in Eq.~\eqref{eq: alignment} cannot invert the relief, as the standard deviation used for scaling is always positive. Compared to the GT hillshade, the DEM-DAV2 hillshade produces a smoother result, as small features and fine-scale rough terrain are attenuated. Overall, the hillshade analysis indicates that DEM-DAV2 can reproduce both the crater morphology and the medium-scale topography.
The improvement of DEM-DAV2 compared to DAV2 is highlighted when conducting the profile analysis in Fig.~\ref{fig:comparisonzeroshotandfinetuned}. We note that the orange profiles, indicating the DEM-DAV2 profiles, yields superior results compared to the green lines, indicating the DAV2 profiles. For example, in the second image in Fig.~\ref{fig:comparisonzeroshotandfinetuned}, DAV2 captures the profile and regional slope more accurately than in the first example, although the hillshade reveals that the craters are still inverted. By contrast, DEM-DAV2 outperforms DAV2 in predicting the GT profile in all three examples, yielding profiles close to the GT. Overall, we note that DEM-DAV2 can be used as a reliable relative lunar height estimator.

\subsection{Quantitative Evaluation}

The quantitative evaluation on the test set is shown in Table~\ref{tab:sota_comparison}. We perform inference on every image in the test set and calculate the metrics per tile. We then take an average over the full test set. We observe, that the DEM-DAV2 yields lower errors on the test dataset compared to DAV2. This highlights the importance of fine-tuning foundation models for remote sensing tasks, given the domain gap. 
The improved evaluation metrics confirm the qualitative observations, showing that fine-tuning the pre-trained DAV2 zero-shot model for DEM estimation yields better results than using it without modification. Note that an analysis of the MAE per test image showed that the error does not increase with the incidence angle (range from 37$^\circ$--70$^\circ$), the emission angle or towards higher latitudes.
\begin{table}[!ht]
\centering
\caption{Quantitative comparison of DAV2 and DEM-DAV2 on the test dataset. The MAE and RMSE columns show the average values for the entire test dataset.}\label{tab:sota_comparison}
\begin{tabular}{l|cc}
\hline
Method & avg. MAE (m) & avg. RMSE (m) \\
\hline
DAV2 & 23.08 & 27.79 \\
DEM-DAV2 & 4.71 & 5.76 \\
\hline
\end{tabular}
\end{table}

\section{Conclusion and Future Work}
In this letter, we introduced a fine-tuning strategy for DAV2 for lunar DEM height estimation. DAV2 for DEM estimation was previously used without fine-tuning in the work of Osadnik et al.~\cite{osadnik2025} and La Grassa et al.~\cite{LAGRASSA2026100118} as a SoTA comparison. However, applying foundation models to the task of remote sensing, poses significant limitations due to a domain gap. We addressed that by inserting new learnable parameters to the model encoder with LoRA and unfroze certain layers in the DPT decoder. For training and evaluation, we utilized the public available SPG-derived DEM products, which are photogrammetric reconstructions with their own error budget rather than measured topography. On our test set, we note an improvement of  $\sim$80\% and  $\sim$79\% in average MAE and RMSE, respectively, indicating that adaptation to lunar terrain mitigates the domain gap inherent in pretrained DAV2. Our results indicate that this enables us to transform DAV2 to a reliable relative height estimator of the lunar surface. Nonetheless, DEM-DAV2 still has some limitations. For example, it cannot be used for real-time DEM estimation due to a rather long inference time on a CPU. The implemented SSI loss function ensures that the predictions remain relative. While it preserves shape up to a scale — supporting shape-based cues such as bowl-shaped impressions — slope analysis and crater morphometry require scale recovery from a low-resolution DEM. This poses a constraint for landing site selection. Also it is only applicable to the lunar surface. Adaptation to other celestial bodies, such as Mars, would need a new training, as the surface is fundamental different. Concerning future research, conducting ablation studies and exploring various training setups for DAV2 could provide valuable insights. Also incorporating DEM-DAV2 as an auxiliary model for crater detection can be very promising. 

\section{Acknowledgements}
This research work is conducted via the Open Space Innovation Platform (\href{https://ideas.esa.int}{https://ideas.esa.int}) as a Co-Sponsored Research Agreement and carried out under the Discovery programme of, and funded by, the European Space Agency (contract number: 4000144734). The view expressed in this publication can in no way be taken to reflect the official opinion of the European Space Agency. This work is also partially funded by the Hochschule Darmstadt and the University of Technology of Troyes within the European University of Technology EUT+.
\bibliographystyle{unsrturl}
\bibliography{references}

\end{document}